\documentclass[sigconf]{acmart}
\usepackage{booktabs}
\usepackage{multirow}
\usepackage{tabularx}
\usepackage{array}
\usepackage{graphicx}
\usepackage{subcaption}
\usepackage{algorithm}
\usepackage{algorithmic}
\usepackage{dsfont}

\usepackage[table]{xcolor} 
\usepackage[dvipsnames]{xcolor}

\AtBeginDocument{%
  }

\copyrightyear{2026}
\acmYear{2026}
\setcopyright{cc}
\setcctype{by}
\acmConference[MM '26]{Proceedings of the 34th ACM International Conference on Multimedia}{November 10--14, 2026}{Rio de Janeiro, Brazil}
\acmBooktitle{Proceedings of the 34th ACM International Conference on Multimedia (MM '26), November 10--14, 2026, Rio de Janeiro, Brazil}
\acmDOI{10.1145/3767308.3835015}
\acmISBN{979-8-4007-2213-4/2026/11}

\begin{document}

\title{Evidence-Grounded Forensic Reasoning for Detecting and
Grounding Multi-Modal Media Manipulation}

\settopmatter{authorsperrow=4}

\author{Yichun Yeh}
\authornote{Both authors contributed equally to the paper.}
\email{yeyijun2024@ia.ac.cn}
\affiliation{%
  \institution{MAIS, CASIA; SAI, UCAS}
  \city{Beijing}
  \country{China}
}

\author{Yiheng Li}
\authornotemark[1]
\email{liyiheng2024@ia.ac.cn}
\affiliation{%
  \institution{SAI, UCAS; MAIS, CASIA}
  \city{Beijing}
  \country{China}
}

\author{Xiaobo Hu}
\email{huxiaobo2024@ia.ac.cn}
\affiliation{%
  \institution{MAIS, CASIA; SAI, UCAS}
  \city{Beijing}
  \country{China}
}

\author{Zhen Lei}
\authornote{Corresponding authors.}
\email{zhen.lei@ia.ac.cn}
\affiliation{%
  \institution{MAIS, CASIA; SAI, UCAS}
  \city{Beijing}
  \country{China}
}

\author{Yang Yang}
\authornotemark[2]
\email{yang.yang@nlpr.ia.ac.cn}
\affiliation{%
  \institution{MAIS, CASIA; SAI, UCAS}
  \city{Beijing}
  \country{China}
}

\renewcommand{\shortauthors}{Yichun Yeh, Yiheng Li, Xiaobo Hu, Zhen Lei, and Yang Yang}

\begin{abstract}
Fake news increasingly relies on cross-modal image-text forgeries, making transparent and verifiable reasoning chains an urgent need for Detecting and Grounding Multi-Modal Media Manipulation (DGM4). Existing methods produce black-box detection results without any decision rationale, limiting their reliability in forensic practice. Multi-modal Large Language Models (MLLMs) offer a natural path toward explainability, but applying them to DGM4 raises two difficulties. First, models tend to generate explanations disconnected from predicted evidence locations, producing unverified attribution. Second, enforcing evidence-conclusion consistency requires active optimization, yet uniform training signals fail to distinguish localization tokens from classification tokens, making multi-head joint training unreliable. We propose a multi-modal manipulation detector based on an Evidence-Grounded Forensic Reasoning (EFR) framework. EFR introduces an Anchor-and-Verify reasoning chain that enforces modality-isolated perception before cross-modal comparison, with conclusion coordinates as explicit anchors to which downstream evidence must spatially correspond. A verifiable reward system then enforces evidence-conclusion consistency during training, while a Modality-Decoupled Advantage (MDA) routing mechanism mitigats credit misassignment across prediction tasks. Experiments show that EFR achieves state-of-the-art performance while producing structured forensic reasoning records that explicitly bind explanations to evidence.
\end{abstract}



\maketitle

\begin{figure}[t]
	\centering
	\includegraphics[width=\columnwidth]{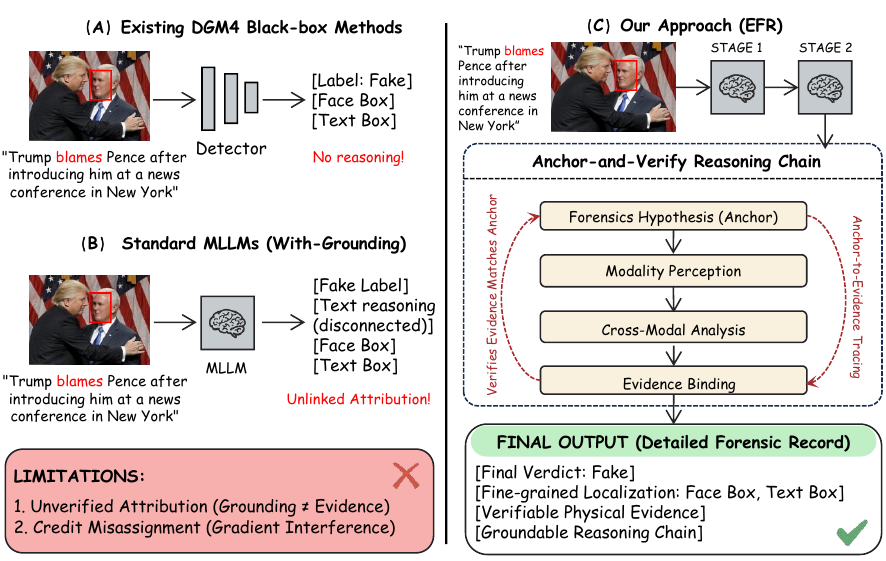}
    \caption{Comparison of existing approaches and our EFR framework. (A) Existing DGM4 methods produce fine-grained results but lack decision rationale. (B) MLLMs offer explanations, yet the reasoning is disconnected from predicted evidence locations, leaving attribution unverified. (C) Our EFR introduces an Anchor-and-Verify reasoning chain that places conclusion coordinates as explicit anchors and requires downstream evidence to spatially match them, enforced by a verifiable reward system during training.}
    \label{fig:intro}
\end{figure}

\section{Introduction}
\label{sec:introduction}
The rapid spread of deepfake technology and large language models~\cite{rossler2018faceforensics, brown2020language} has made multi-modal misinformation increasingly easy to produce and hard to detect~\cite{pei2024deepfake}. Isolated face manipulation~\cite{zhao2021multi} and textual fabrication~\cite{nan2021mdfend} have expanded into coordinated cross-modal forgeries where images and their associated texts are manipulated simultaneously, posing escalating threats to the credibility of public media. To address these threats, the DGM4 task~\cite{shao2023detecting} requires models to jointly perform authenticity classification, manipulation type identification, and fine-grained localization of manipulated faces and text tokens, a combination that far exceeds the complexity of single-modal detection. Existing DGM4 methods follow two main directions: cross-modal feature alignment and semantic interaction~\cite{zhang2025asap, li2025unleashing}, and fine-grained localization through richer supervision such as frequency-domain features and multi-scale objectives~\cite{liu2025unified, liu2025idseq, yu2025fine}. Despite steady progress, they treat detection as a closed prediction problem, outputting labels and bounding boxes without any decision rationale. Yet real-world forensic workflows such as editorial fact-checking and legal content authentication demand decisions that can be audited and challenged. These limitations call for a detection paradigm that produces structured, verifiable reasoning alongside its predictions.

Beyond these task-specific methods, MLLMs~\cite{wang2024qwen2, liu2023visual, liu2024improved} have recently been applied to single-modality manipulation detection with localization and explanation~\cite{he2025vlforgery,kundu2025truthlens,liu2024forgerygpt,xu2026mare}. Leveraging cross-modal understanding and language generation, these approaches~\cite{xu2024fakeshield, huang2025sida,lin2025seeing} offer a natural path toward explainable detection, with recent efforts adding pattern-aware reasoning, self-reflection, and two-stage training~\cite{tan2025veritas, gao2025fakereasoning, jiang2025tridf, shen2025df}. Rather than pursuing generic reasoning, we identify a critical yet overlooked flaw shared by these methods: their explanations are never tied to the model's own predicted localization, so their interpretability remains superficial. This motivates a new direction, and applying MLLMs to DGM4 further exposes two problems amplified by its cross-modal complexity. First, \textit{unverified attribution}: without structural constraints, MLLMs produce coherent-looking reasoning yet cannot verify that cited evidence matches the predicted conclusion coordinates~\cite{zhu2026analyzing, zhu2024unraveling}. Second, \textit{credit misassignment}: DGM4 packs heterogeneous heads (classification, face localization, and text localization) into one output sequence, so a uniform signal over all tokens lets gradients from one head interfere with others, destabilizing joint training.

To address both problems, we propose the Evidence-Grounded Forensic Reasoning (EFR) framework. Since existing benchmarks provide no forensic rationale and contain labeling noise, we first construct a high-quality reasoning dataset as the training foundation. EFR then operates in two coordinated stages. The first stage introduces an Anchor-and-Verify reasoning chain that organizes outputs into modality-isolated perception, cross-modal conflict analysis, and evidence binding, placing conclusion coordinates as explicit anchors that downstream evidence must match, thereby closing the attribution gap. The second stage then optimizes the model with reinforcement learning: a five-component verifiable reward system reinforces this coordinate-level consistency, while a Modality-Decoupled Advantage (MDA) routing mechanism sends task-specific advantages to their corresponding tokens, eliminating the cross-head interference. Our main contributions are as follows:

\begin{itemize}
\item We propose EFR, an evidence-grounded forensic reasoning framework for DGM4 that, unlike black-box detectors and unconstrained MLLM explanations, grounds every forensic conclusion in a spatially verifiable location through anchor-first verifiable binding.

\item We introduce an Anchor-and-Verify reasoning chain that constrains cited evidence to match the predicted anchors, together with a Modality-Decoupled Advantage routing scheme that delivers task-specific signals to their corresponding prediction tokens; both are trained through a two-stage pipeline over a curated 50K forensic reasoning dataset.

\item Extensive experiments on DGM4 show that EFR attains state-of-the-art detection and competitive grounding, while producing structured reasoning records whose cited evidence is spatially consistent with the predicted anchors.
\end{itemize}  

\begin{figure*}[t]
	\centering
	\includegraphics[width=\textwidth]{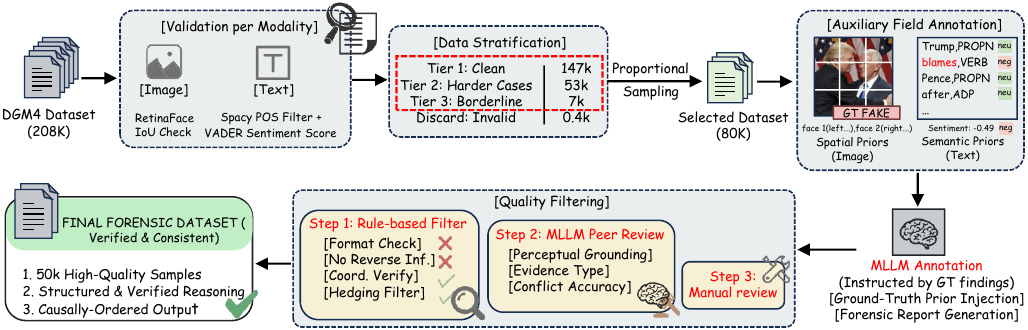}
	\caption{The forensic reasoning annotation data construction pipeline. Raw DGM4 samples (208K) are filtered and stratified by quality, with spatial and semantic priors extracted to guide MLLM annotation. A three-step quality filtering stage combining rule-based checks, MLLM peer review, and manual inspection yields 50K verified forensic reasoning samples.\label{fig:data}}
\end{figure*}

\section{Related Works}

\noindent \textbf{Multi-Modal Manipulation Detection.} 
Early efforts addressed coarse-grained binary classification for fake-news and out-of-context misinformation detection, treating an image-text pair as a single real-or-fake decision. HAMMER~\cite{shao2023detecting} reframes the problem as DGM4, which jointly detects manipulation, classifies its fine-grained type, and grounds the tampered image regions and text tokens. Subsequent work advances the task along two directions. One strengthens cross-modal alignment so that subtle semantic inconsistencies become separable: ASAP~\cite{zhang2025asap} adapts prompts to the input pair, while CSCL~\cite{li2025unleashing} enforces consistency-based objectives across modalities. The other sharpens localization through richer supervision, as in UFAFormer~\cite{liu2025unified} with frequency-domain cues, IDseq~\cite{liu2025idseq} with decoupled grounding, and FMSN~\cite{yu2025fine} with multi-scale signals. Beyond these, RamDG~\cite{shen2025beyond} targets coordinated semantic manipulations, and LADLE-MM~\cite{cardullo2025ladle} and CIEC~\cite{yu2026ciec} tackle limited-annotation settings, while DGM4+~\cite{singh2025dgm4+} and MDSM~\cite{zhang2025coherence} extend the benchmark with global scene inconsistencies and MLLM-crafted narratives. Despite steady progress, these methods remain prediction-oriented, producing bounding boxes and labels without any rationale behind their decisions.

\noindent \textbf{MLLMs for Forgery Detection.} The strong cross-modal reasoning of MLLMs has shifted forgery detection toward explainability, moving beyond a binary verdict to a human-readable justification. Early work~\cite{jia2024can} probes their zero-shot capability through prompt engineering, and later methods~\cite{chen2025mgffd, huang2025sida, xu2024fakeshield, qin2025fake} tie fine-grained tampered regions to textual explanations via multi-granularity prompts. To make the reasoning process explicit, AIGI-Holmes~\cite{zhou2025aigi} and FakeReasoning~\cite{gao2025fakereasoning} impose chain-of-thought formats aligned with forgery attributes, while VERITAS~\cite{tan2025veritas} and RAIDX~\cite{li2025raidx} improve generalization through pattern-aware reasoning and retrieval-augmented evidence. More recently, DF-LLaVA~\cite{shen2025df} injects external knowledge with conflict-driven self-reflection for single-image detection, and TriDF~\cite{jiang2025tridf} evaluates perception and hallucination to gauge interpretability. Yet all of these generate explanations without enforcing consistency between the cited evidence and the predicted conclusion coordinates, so a plausible-sounding rationale may still point elsewhere, leaving attribution unverified; applied to the multi-head DGM4 task, they further suffer credit misassignment under a uniform training signal. In contrast, EFR enforces anchor-first, coordinate-level verifiable binding and decouples per-task optimization, addressing both limitations.

\noindent \textbf{Reinforcement Learning for LLMs.} RL-based post-training has evolved from PPO-based RLHF toward scalable reasoning optimization~\cite{schulman2017proximal, rafailov2023direct, ouyang2022training}. GRPO~\cite{shao2024deepseekmath} removes the explicit value model, estimating advantages from group-level reward comparisons. Later variants refine this scheme: DAPO~\cite{yu2025dapo} counters entropy collapse and training instability, SRPO~\cite{zhang2025srpo} scales via a two-stage curriculum, GSPO~\cite{zheng2025group} stabilizes updates with sequence-level importance ratios, and GDPO~\cite{liu2026gdpo} decouples reward normalization for multi-reward settings. All of them, however, apply the reward uniformly across output tokens, ignoring the semantic boundaries of structured multi-modal outputs. Our MDA routes task-specific advantages along these boundaries within the reasoning chain, mitigating credit misassignment in multi-head training.

\begin{figure*}[t]
	\centering
	\includegraphics[width=\textwidth]{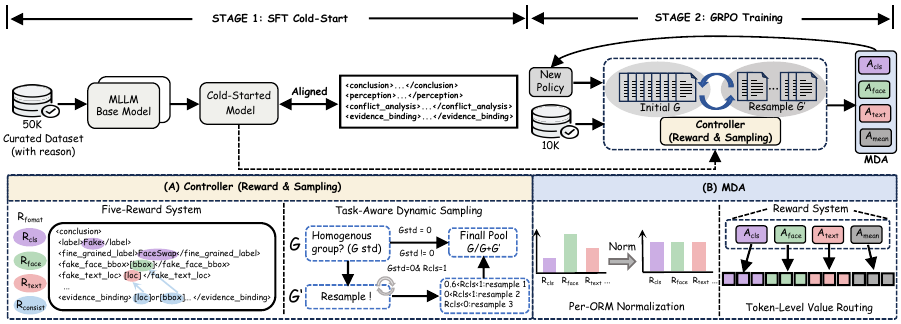}
	\caption{Overview of the two-stage EFR training framework. Stage 1 fine-tunes a base MLLM on 50K curated forensic reasoning samples to initialize structured output following the Anchor-and-Verify format. Stage 2 optimizes the cold-started model via GRPO. (A) A five-component verifiable reward system combined with task-aware dynamic resampling maintains training diversity. (B) MDA performs per-ORM normalization across modality-specific reward heads and routes task-specific advantage values to their corresponding prediction tokens, mitigating credit misassignment.}
	\label{fig:method}
\end{figure*}

\section{Methodology}
\subsection{Overview}

Directly applying MLLMs to DGM4 exposes two fundamental problems:
\textit{unverified attribution}, where explanations become disconnected from predicted manipulation locations, and \textit{credit misassignment}, where joint optimization across heterogeneous prediction heads suffers under uniform advantage estimation. As illustrated in Figure~\ref{fig:method}, EFR addresses both through two coordinated stages. Stage 1 (Sec.~\ref{sec:stage1}) tackles unverified attribution with a structured reasoning chain that binds forensic conclusions to spatially grounded evidence, internalized via supervised fine-tuning. Stage 2 (Sec.~\ref{sec:stage2}) mitigates credit misassignment by replacing uniform advantage estimation with task-specific signals routed to their corresponding prediction tokens.

\subsection{Structured Reasoning Cold-Start }
\label{sec:stage1}

To equip the model with structured forensic reasoning capability, we construct a high-quality forensic reasoning dataset from the DGM4 benchmark and use it to initialize the model via supervised fine-tuning to internalize the Anchor-and-Verify structure.

\subsubsection{Anchor-and-Verify Reasoning Chain}
\label{sec:cot}

Without structural constraints, MLLMs produce reasoning chains that appear coherent but lack a mechanism to verify that cited evidence corresponds to predicted conclusion coordinates. We address this by designing the Anchor-and-Verify reasoning chain, a structured chain-of-thought format that organizes model outputs into four ordered stages, each serving a distinct forensic function, as illustrated in Figure~\ref{fig:intro}.

\noindent \textbf{Forensic Hypothesis.} All predicted outputs are placed at the beginning of the sequence as a Forensic Hypothesis Block. The predicted coordinates serve as explicit anchors that all downstream stages must spatially correspond to, establishing a verifiable reference before any explanation is generated.

\noindent \textbf{Modality-Isolated Perception.} Independent analysis is enforced through two strictly separated sub-modules. The visual module describes only physically observable image features, prohibited from referencing caption content or making authenticity judgments; the text module independently analyzes caption semantics and logical structure without referencing the image. This separation reduces the influence of language priors on visual analysis.

\noindent \textbf{Cross-Modal Conflict Analysis.} The two perceptual outputs are explicitly compared and their disagreement is graded into two levels: strong conflicts, where the image and text assert mutually exclusive facts, and weak mismatches, where the two modalities are thematically related but neither corroborates nor contradicts the other. This graded assessment produces fine-grained, evidence-grounded conflict judgments instead of a coarse binary decision.

\noindent \textbf{Physical Evidence Binding.} Conflict analysis is grounded in concrete physical locations. Let $\hat{b}$ and $\hat{s}$ denote the predicted face bounding box and text token span from the Forensic Hypothesis Block. Visual evidence must cite a bounding box $\hat{b}'$ with verifiable observations, and text evidence must cite a token span $\hat{s}'$ with an explanation of the semantic discontinuity. Both must satisfy $\hat{b}' \approx \hat{b}$ and $\hat{s}' \subseteq \hat{s}$, ensuring every forensic conclusion is traceable to a specific location. This constraint is enforced by the verifiable reward system during reinforcement learning (Sec.~\ref{sec:reward}).

\subsubsection{Forensic Reasoning Data Construction}
\label{sec:data}

With the Anchor-and-Verify chain established as the target structure, we construct the training data needed to instantiate it. The DGM4 dataset provides manipulation labels and localization annotations but no forensic rationale, and its 208K samples include substantial labeling noise. We design a systematic pipeline to clean, stratify, annotate, and filter the data, as shown in Figure~\ref{fig:data}.

\noindent \textbf{Data cleaning and stratification.} Samples are first validated per modality for face-localization and text-modification quality (details in the supplementary). Based on these quality dimensions and manual review of 6K critical samples, the dataset is stratified into three tiers: Tier 1 (147K clean), Tier 2 (53K harder multi-face or complex-token cases), and Tier 3 (7K borderline); invalid samples (0.4K) are discarded. Proportional sampling over Tier 1 and Tier 2 yields an 80K dataset for annotation.

\noindent \textbf{Auxiliary field annotation.} Before reasoning generation, each sample is enriched with structured spatial and semantic priors: per-face position and area-ranked indices with an explicit \texttt{<FAKE>} tag on the ground-truth fake face for images, and a token map with polarity scores for eligible candidate tokens for text. These priors give the generation model unambiguous references that mitigate positional confusion in multi-face scenes and ground text localization in precise token structure.

\noindent \textbf{Reasoning generation and quality filtering.} To generate reasoning chains, we condition a large-scale MLLM on the ground-truth findings, so that it reports physically consistent evidence for a known verdict rather than inferring the verdict itself. Each generated chain then passes three filters: rule-based checks on format and coordinate references, an independent MLLM review of perceptual grounding and conflict accuracy, and manual inspection of flagged cases. This yields the final 50K forensic reasoning samples.

\subsubsection{Structured Reasoning Cold-Start}
\label{sec:sft}

Direct RL training on the Anchor-and-Verify format is unstable, as the complex structured output requires the model to produce parseable multi-block responses before any reward signal can be meaningfully applied. We therefore first perform supervised fine-tuning (SFT) on our curated 50K forensic dataset $\mathcal{D}$, establishing a reliable reference policy $\pi_{\mathrm{ref}}$ that can consistently adhere to the formatting constraints. Following standard instruction-tuning paradigms~\cite{ouyang2022training}, our optimization objective is to minimize the autoregressive cross-entropy loss:
\begin{equation}
    \mathcal{L}_{\text{SFT}}(\theta) = - \mathbb{E}_{(x,\, y) \sim \mathcal{D}} 
    \left[ \sum_{t=1}^{L} \log \pi_\theta(y_t \mid x, y_{<t}) \right],
    \label{eq:sft}
\end{equation}
where $x$ represents the multi-modal prompt, $y = (y_1, \ldots, y_L)$ denotes the target sequence of length $L$ comprising the complete Anchor-and-Verify reasoning chain, and $\pi_\theta$ is the policy parameterized by $\theta$. Through this phase, the model internalizes the Anchor-and-Verify reasoning structure and learns to produce physically consistent observations and cross-modal evidence, providing a robust initialization for the subsequent RL stage.

\subsection{Modality-Decoupled Policy Optimization}
\label{sec:stage2}

While cold-start SFT equips the model with format adherence, supervised imitation provides no explicit feedback on spatial consistency between conclusion anchors and cited evidence, nor does it resolve credit misassignment across multi-head outputs. We therefore apply reinforcement learning on top of the cold-start model $\pi_{\mathrm{ref}}$, building on Group Relative Policy Optimization (GRPO)~\cite{shao2024deepseekmath}, which estimates policy gradients from group-level reward comparisons without a separate value network. For each prompt $x$, a group of $G$ completions $\{y_i\}_{i=1}^{G}$ is sampled from the current policy $\pi_\theta$, and the objective is:

\begin{equation}
\begin{aligned}
\mathcal{L}_{\text{GRPO}} &= -\mathbb{E}\Bigg[ \frac{1}{G}\sum_{i=1}^{G}\frac{1}{T_i} \sum_{t=1}^{T_i} \min\Big(\rho_{i,t}A_i, \\
&\quad \mathrm{clip}(\rho_{i,t}, 1-\varepsilon, 1+\varepsilon)A_i\Big) - \beta\mathbb{D}_{\text{KL}} \left[\pi_\theta\|\pi_{\mathrm{ref}}\right] \Bigg],
\end{aligned}
\label{eq:grpo}
\end{equation}

where $T_i$ is the token length of completion $i$; $\varepsilon$ is the clipping threshold that limits the update of the policy per-step; and $\beta$ is the KL penalty coefficient that regularizes $\pi_\theta$ with the reference policy for cold-start $\pi_{\mathrm{ref}}$. The importance sampling ratio $\rho_{i,t}$ and the group-normalized advantage $A_i$ are:

\begin{equation}
\rho_{i,t} = \frac{\pi_\theta(y_{i,t}\mid y_{i,<t},\,x)}
{\pi_{\text{ref}}(y_{i,t}\mid y_{i,<t},\,x)},
\label{eq:ratio}
\end{equation}

\begin{equation}
A_i = \frac{R_i - \mu_G}{\sigma_G + \epsilon},\quad
\mu_G = \frac{1}{G}\sum_{i=1}^{G}R_i,\quad
\sigma_G = \sqrt{\frac{1}{G}\sum_{i=1}^{G}(R_i-\mu_G)^2},
\label{eq:adv}
\end{equation}

where $R_i$ is the total scalar reward for completion $i$ and $\epsilon$ is a small 
constant for numerical stability.

Applying GRPO directly to DGM4, however, is problematic: aggregating heterogeneous prediction targets operating on different reward scales into a single $R_i$ produces noisy advantage estimates, and the uniform scalar advantage $A_i$ applied identically to all tokens conflates gradients across classification, localization, and reasoning tokens. We address both with a fine-grained verifiable reward system and an MDA routing mechanism.

\subsubsection{Verifiable Reward System}
\label{sec:reward}

Designing effective rewards for DGM4 requires covering output format, classification accuracy, multi-modal localization, and reasoning consistency in a single framework. Inspired by verifiable reward design in mathematical reasoning~\cite{shao2024deepseekmath, guo2025deepseek}, we define five independent, deterministic rule-based Outcome Reward Models (ORMs), which assess our structured outputs precisely while keeping training signals stable and reproducible. To decouple signals across tasks, samples are partitioned by manipulation type (\textit{face}, \textit{text}, \textit{compound}, \textit{orig}); inapplicable ORMs are excluded from the per-sample sum, so gradients flow only through relevant heads.

\noindent \textbf{Format Compliance ($R_{\text{format}}$)} gates the reward: it enforces XML structural completeness and penalizes placeholder copying, ensuring outputs are parseable before other ORMs score them.

\noindent \textbf{Hierarchical Classification ($R_{\text{cls}}$)} jointly evaluates binary authenticity and fine-grained manipulation type with a cost-sensitive design, where missed detections are penalized more heavily than false positives. For valid predictions, fine-grained reward follows a hierarchical partial-order design:
\begin{equation}
R_{\text{cls}} = r_{\text{base}}
    + w_{\text{fam}} \cdot F1_{\text{fam}}
    + w_{\text{sub}} \cdot r_{\text{sub}}
    + w_{\text{exact}} \cdot \mathds{1}[\hat{y}=y],
\label{eq:r1}
\end{equation}
where $r_{\text{base}}$ is a base score, $F1_{\text{fam}}$ measures family-level overlap (e.g., face vs.\ text family), $r_{\text{sub}}$ captures within-family subtype accuracy, and $w_{\text{fam}}, w_{\text{sub}}, w_{\text{exact}}$ are fixed weights reflecting the taxonomy.

\noindent \textbf{Localization ($R_{\text{face}}$, $R_{\text{text}}$)} score spatial grounding for the relevant categories: $R_{\text{face}}$ maps predicted-box IoU to a tiered reward, and $R_{\text{text}}$ measures token-span overlap via Jaccard similarity with subtype-specific curves.

\noindent \textbf{Consistency ($R_{\text{consistency}}$)} verifies internal coherence of the reasoning chain, rewarding cases where the evidence cited in the \texttt{evidence\_binding} block spatially and semantically matches the conclusion anchors and where evidence types match the predicted category, thereby instantiating the physical binding constraints of Sec.~\ref{sec:cot}. The full decomposition is given in the supplementary.

The five components are integrated into a final scalar reward for each response:
\begin{equation}
    R = \lambda_f R_{\text{format}} + \lambda_c R_{\text{cls}} + 
    \lambda_v R_{\text{face}} + \lambda_t R_{\text{text}} + 
    \lambda_r R_{\text{consistency}},
    \label{eq:reward}
\end{equation}
where $\lambda_f, \lambda_c, \lambda_v, \lambda_t, \lambda_r$ balance the contribution of each component. While these ORMs provide task-specific signals, translating them into precise gradient updates requires routing each signal to its corresponding prediction tokens, addressed by MDA next.

\begin{table*}[!t] 
\renewcommand{\arraystretch}{1.10} 
\setlength{\tabcolsep}{1.5mm} 
\begin{center}
\caption{\textbf{Comparison of state-of-the-art methods on DGM4.}$\downarrow$ means lower is better. \textbf{Bold}/\underline{underline}: best/second best in each group. $\dagger$ denotes zero-shot evaluation without fine-tuning. $\ddagger$For MLLM-based methods, mAP is reported as $a/b$: from token log probabilities and from the hard $0/1$ prediction, respectively (Sec.~\ref{sec:Setup}).}
\label{tab1}
\resizebox{\textwidth}{!}{  
\begin{tabular}{ l c c c c c c c c c c c c }   
\hline
\multirow{2}{*}{Method}
&\multicolumn{3}{c}{Binary Cls}&\multicolumn{3}{c}{Multi-label Cls}&\multicolumn{3}{c}{Image Grounding}&\multicolumn{3}{c}{Text Grounding}\\  
&AUC&EER$\downarrow$&ACC&mAP$^\ddagger$&CF1&OF1&IoU$_m$&IoU$_{50}$&IoU$_{75}$&PR.&RE.&F1 \\
\hline
HAMMER~\cite{shao2023detecting} \small{(CVPR'23)}&$93.19$&$14.10$&$86.39$&$86.22$&$79.37$&$80.37$&$76.45$&$83.75$&$76.06$&$75.01$&$68.02$&$71.35$\\
HAMMER++~\cite{shao2024detecting} \small{(TPAMI'24)}&$93.33$&$14.06$&$86.66$&$86.41$&$79.73$&$80.71$&$76.46$&$83.77$&$76.03$&$73.05$&$72.14$&$72.59$\\
ViKI~\cite{li2024towards} \small{(IF'24)}&$93.51$&$13.87$&$86.67$&$86.58$&$81.07$&$80.10$&$76.51$&$83.95$&$75.77$&$77.79$&$66.06$&$73.44$\\
UFAFormer~\cite{liu2025unified} \small{(IJCV'24)}&$93.81$&$13.60$&$86.80$&$87.85$&$80.31$&$81.48$&$78.33$&$85.39$&$79.20$&$73.35$&$70.73$&$72.02$\\
MSF~\cite{wang2024exploiting} \small{(ICASSP'24)}&$95.11$&$11.36$&$88.75$&$\mathbf{91.42}$&$83.60$&$84.38$&$80.83$&$88.35$&$80.39$&$76.51$&$70.61$&$73.44$\\
IDseq~\cite{liu2025idseq} \small{(AAAI'25)}&$94.55$&$11.40$&$\underline{88.94}$&$90.01$&$83.00$&$84.90$&$\underline{83.33}$&$\underline{89.39}$&$\mathbf{86.10}$&$75.96$&$71.23$&$73.52$\\
ASAP~\cite{zhang2025asap} \small{(CVPR'25)}&$94.38$&$12.73$&$87.71$&$88.53$&$81.72$&$82.89$&$77.35$&$84.75$&$76.54$&$\mathbf{79.38}$&$73.86$&$\mathbf{76.52}$\\
\hline
Qwen3-VL-8B$^\dagger$~\cite{bai2025qwen3}& 64.42 & 40.06 & 54.71 & $30.34/26.48$ & 0.02 & 0.02 & 32.65 & 32.54 & 32.54 & 21.66 & 1.59 & 2.97\\
InternVL3.5-8B$^\dagger$~\cite{wang2025internvl3}& 50.00 & 50.00 & 33.28 & $28.26/20.76$ & 0.00 & 0.00 & 13.00 & 12.60 & 12.55 & 14.78 & 0.84 & 1.59\\
\hline
\textsc{EFR} (cold-start)& $\underline{96.87}$ & $\underline{9.38}$ & $90.65$ & $89.60/81.50$ & $\underline{86.94}$ & $\underline{86.68}$ & $83.22$ & $89.24$ & $85.21$ & $\underline{78.68}$ & $\underline{72.14}$ & $75.27$ \\
\rowcolor{Gray!20}
\textsc{EFR} (ours)& $\mathbf{96.97}$ & $\mathbf{9.18}$ & $\mathbf{90.82}$ & $\underline{90.41}/81.58$ & $\mathbf{87.19}$ & $\mathbf{87.05}$ & $\mathbf{83.49}$ & $\mathbf{89.41}$ & $\underline{85.43}$ & $76.00$ & $\mathbf{75.61}$ & $\underline{75.80}$\\
\hline
\end{tabular}
}
\end{center}
\end{table*}

\subsubsection{Modality-Decoupled Advantage Routing}
\label{sec:mda}
The verifiable reward system provides fine-grained signals across five task dimensions; MDA routes per-task advantages to their corresponding tokens to turn these into precise gradient updates 
(Figure~\ref{fig:method}).

\noindent \textbf{Per-ORM Normalization.} Since $R_{\text{format}}$--$R_{\text{consistency}}$ operate on different ranges and distributions, aggregating them into a single advantage distorts the relative contribution of each task. MDA normalizes each ORM independently within the group:
\begin{equation}
\hat{r}_k^{(i)} = \frac{r_k^{(i)} - \mu_k}{\sigma_k + \epsilon},
\label{eq:norm}
\end{equation}
where $r_k^{(i)}$ is the reward from ORM $k$ for completion $i$, and $\mu_k$, $\sigma_k$ are the group mean and standard deviation of that ORM. Task-specific advantages are then derived: $A_{\text{cls}}$ from $\hat{r}_{\text{cls}}$; $A_{\text{face}}$ from $\hat{r}_{\text{face}}$ and the face-relevant portion of $\hat{r}_{\text{consistency}}$; and $A_{\text{text}}$ from $\hat{r}_{\text{text}}$ and the text-relevant portion of $\hat{r}_{\text{consistency}}$.

\noindent \textbf{Token-Level Routing.} This routing is made possible by the structured XML boundaries established in Sec.~\ref{sec:cot}: the Anchor-and-Verify format provides explicit block-level semantic boundaries that map each output token to a specific prediction task. MDA parses these boundaries at training time and assigns token-level advantages as:
\begin{equation}
A_t = \begin{cases}
    \displaystyle\sum_{k \in \mathcal{K}} A_k \cdot \mathds{1}[t \in \mathcal{T}_k] 
        & \text{if } t \in \bigcup_k \mathcal{T}_k \\[6pt]
    \bar{A} & \text{otherwise}
\end{cases}
, \mathcal{K} = \{\text{cls}, \text{face}, \text{text}\},
\label{eq:routing}
\end{equation}
where $\mathcal{T}_k$ is the set of token positions belonging to task $k$, and $\bar{A}$ is the mean advantage across all tasks, applied to shared reasoning-chain tokens that do not belong to any specific prediction block. Task-specific tokens receive the advantage of their corresponding head; reasoning-chain tokens receive $\bar{A}$, reflecting their shared role across all prediction tasks. As a global parseability gate, $R_{\text{format}}$ is not routed to any single block but scales the whole-sequence reward.

Following DAPO~\cite{yu2025dapo}, we exclude the KL penalty term 
from the objective (i.e., $\beta D_{\mathrm{KL}}[\pi_\theta \| \pi_{\mathrm{ref}}]$), 
as the model distribution is expected to diverge substantially from 
the cold-start reference during forensic reasoning, making this 
constraint unnecessary and potentially harmful to exploration. 
The final objective replaces the uniform scalar advantage 
in Eq.~\ref{eq:grpo} with the token-level routed advantage:
\begin{equation}
\begin{aligned}
\mathcal{L}_{\text{MDA}} &= -\mathbb{E}\Bigg[ \frac{1}{G}\sum_{i=1}^{G}\frac{1}{T_i} \sum_{t=1}^{T_i} \min\!\Big(A_t^{(i)}\rho_{i,t}, \\
&\qquad A_t^{(i)}\,\mathrm{clip}(\rho_{i,t}, 1{-}\varepsilon, 1{+}\varepsilon)\Big) \Bigg],
\end{aligned}
\label{eq:mda}
\end{equation}
\noindent \textbf{Task-Aware Dynamic Resampling.} Effective token-level routing requires stable and informative advantage estimates for each task dimension. DGM4's heterogeneous task structure makes this non-trivial: original samples activate only a subset of ORMs and tend to produce low reward variance within a group, causing advantage estimates to degenerate. To stabilize training, MDA adopts a task-aware dynamic resampling strategy. Groups meeting task-specific quality thresholds across all active ORMs are exempted; for others, additional completions are sampled with intensity proportional to per-ORM error severity, ensuring every prediction head receives informative gradient signals.

By integrating per-ORM normalization, token-level routing, and task-aware resampling, MDA ensures that each prediction head receives optimization signals derived only from task-relevant reward dimensions. This mechanism is made possible by the explicit structural boundaries of the Anchor-and-Verify reasoning chain. Finally, the reasoning format and the optimization mechanism are tightly co-designed to be mutually reinforcing.

\section{Experiments}

\subsection{Setup}
\label{sec:Setup}

\noindent \textbf{Datasets.}
We evaluate EFR on the DGM4 dataset~\cite{shao2023detecting}, a large-scale benchmark for detecting and grounding multi-modal media manipulation in the news domain. It comprises 230,000 image-text pairs (77,426 genuine and 152,574 manipulated) from four major outlets (The Guardian, BBC, USA TODAY, and The Washington Post), with manipulations spanning \textit{face swap}, \textit{face attribute}, \textit{text swap}, and \textit{text attribute}, covering both visual and textual forgery.

\noindent \textbf{Evaluation Metrics.}
Following prior works~\cite{shao2023detecting,shao2024detecting,liu2025idseq,li2025unleashing,zhang2025asap}, we adopt a comprehensive metric suite covering both coarse-grained detection and fine-grained grounding. Binary detection is assessed by AUC, EER ($\downarrow$), and ACC; multi-label classification by mAP, CF1, and OF1. Image grounding is measured by IoU$_m$, IoU$_{50}$, and IoU$_{75}$; text grounding by token-level Precision (PR), Recall (RE), and F1. 

As EFR generates predictions autoregressively, AUC is computed from the token-level log probabilities of the predicted authenticity label. Fine-grained labels are discrete and provide no continuous confidence score. For a fair comparison with prior work, we recover the per-class confidences from the token-level log probabilities at the fine-grained-label position and evaluate mAP under the standard protocol; we additionally report mAP on the hard $0/1$ prediction, which is used in our ablations.

\noindent \textbf{Implementation Details.}
We build EFR on Qwen3-VL-7B~\cite{bai2025qwen3}, and use Qwen3-VL-32B~\cite{bai2025qwen3} as the annotation model for forensic reasoning data construction, generating evidence reports conditioned on ground-truth findings via confirmed-findings injection. Training proceeds in two stages~\cite{guo2025deepseek,shao2024deepseekmath,yu2025dapo}, both using LoRA~\cite{hu2022lora} ($r{=}128$, $\alpha{=}256$). Stage~1 performs cold-start SFT on 50K forensic reasoning samples for 3 epochs (learning rate $5{\times}10^{-5}$, batch size 64) to initialize the Anchor-and-Verify structure. Stage~2 applies GRPO~\cite{shao2024deepseekmath} on 10K samples for 1 epoch (learning rate $1{\times}10^{-6}$, batch size 16, group size $G{=}8$, temperature 1.0). Following DAPO~\cite{yu2025dapo}, both KL coefficients $\beta$ and $\beta'$ are set to 0, as the policy is expected to diverge substantially from the cold-start reference during optimization.

\subsection{Results}

Table~\ref{tab1} compares EFR against dedicated DGM4 detectors and general-purpose MLLMs. EFR achieves state-of-the-art binary detection and multi-label F1, remains competitive on mAP and grounding.

\noindent \textbf{Comparison with State-of-the-Art Methods.}
The advantage of EFR is concentrated on fine-grained attribution rather than detection. Binary detection is close to saturation for all recent methods, and EFR advances it by a modest 1.86 AUC over the strongest baseline. In multi-label classification the picture changes: CF1 and OF1 improve by 3.59 and 2.15 points, roughly twice the margin. Determining whether an item has been manipulated reduces to a low-level consistency check, whereas identifying which of the four manipulations occurred requires each hypothesis to be evaluated against its own evidence, the procedure the Anchor-and-Verify chain makes explicit. The grounding results are consistent with this account. EFR obtains the highest IoU$_m$ and IoU$_{50}$ despite emitting coordinates as text tokens, and falls behind IDseq only at the strictest threshold, where a dedicated localization head retains sub-pixel precision beyond the resolution of a language decoder. mAP is the only metric where a discriminative head stays ahead, and the cause is protocol, not capability: prior methods read mAP off continuous multi-label scores, while ours comes from token log probabilities (Sec.~\ref{sec:Setup}), a coarser signal that hurts ranking but not the decisions, as CF1 and OF1 confirm.

\noindent \textbf{Effect of Stage 2 Optimization.}
Stage 2 does not make the model uniformly better; it moves the model to a different operating point. Detection barely changes, as expected from a metric already above 96.8 after cold-start. What changes is text grounding, which after cold-start is strongly precision-skewed: the model reports only the spans it is certain about and misses the rest. Stage 2 trades 2.68 points of precision for 3.47 points of recall, turning an asymmetric operating point into a balanced one and improving F1. This is the intended effect of the reward design. Static annotations supervise the exact span and provide no signal that separates a partially recovered span from a missed one, so supervised training has no reason to extend a confident partial prediction; the verifiable rewards score coverage directly, and the model responds by widening its evidence.

\noindent\textbf{Zero-shot MLLMs.}
General-purpose MLLMs can rank but cannot decide. Qwen3-VL-8B and InternVL3.5-8B obtain mAP of 30.34 and 28.26, close to what the class priors alone would yield, yet their CF1 and OF1 collapse to near zero: the models occasionally place manipulated items above authentic ones, but almost never emit the correct label. Without task-specific training, visual-linguistic competence does not translate into forensic attribution.

\begin{table}[!t]
\renewcommand{\arraystretch}{1.10}
\setlength{\tabcolsep}{4pt}
\begin{center}
\caption{\textbf{Ablation on MDA routing components.} Each row adds one component to the previous. \textbf{Bold}/\underline{underline}: best/second best. mAP is the hard $0/1$ variant (Sec.~\ref{sec:Setup}).}
\label{tab:ablation_method}
\resizebox{\columnwidth}{!}{
\begin{tabular}{l cccc}
\hline
\textbf{Method} & \textbf{ACC} & \textbf{mAP} & \textbf{IoU$_m$} & \textbf{F1} \\
\hline
SFT + GRPO                                         & 90.74          & 78.88          & 82.68          & 75.16          \\
~~+ Task-Aware Dyn. Resampling                    & \textbf{90.86} & \underline{80.26} & 82.50          & \underline{75.69} \\
~~+ Per-ORM Normalization                         & 90.41          & 77.37          & \underline{82.78} & 74.17          \\
~~+ Token-Level Routing \small{(\textit{Ours})}             & \underline{90.82} & \textbf{81.58} & \textbf{83.49} & \textbf{75.80} \\
\hline
$\Delta$ vs.\ SFT + GRPO & {\color{red}+0.08} & {\color{red}+2.70} & {\color{red}+0.81} & {\color{red}+0.64} \\
\hline
\end{tabular}}
\end{center}
\vspace{-1em}
\end{table}

\begin{table}[!t]
\renewcommand{\arraystretch}{1.10}
\setlength{\tabcolsep}{4pt}
\begin{center}
\caption{\textbf{Ablation on reward components in MDPO.} \checkmark denotes an active reward. \textbf{Bold}/\underline{underline}: best/second best. mAP is the hard $0/1$ variant (Sec.~\ref{sec:Setup}).}
\label{tab:ablation_reward}
\resizebox{\columnwidth}{!}{
\begin{tabular}{ccccc c c c c}
\hline
\textbf{$R_{\text{fmt}}$} & \textbf{$R_{\text{acc}}$} & \textbf{$R_{\text{bbox}}$} & \textbf{$R_{\text{text}}$} & \textbf{$R_{\text{consist}}$} & \textbf{ACC} & \textbf{mAP} & \textbf{IoU$_m$} & \textbf{F1} \\
\hline
\checkmark & \checkmark &            &            &            & \textbf{90.84} & 77.21 & 82.13 & 75.49 \\
\checkmark & \checkmark & \checkmark &            &            & 90.78 & 81.35 & 83.26 & 75.80 \\
\checkmark & \checkmark &            & \checkmark &            & \underline{90.83} & 81.08 & 81.83 & 75.60 \\
\checkmark & \checkmark & \checkmark & \checkmark &            & 90.78 & \textbf{81.59} & \underline{83.42} & \textbf{75.89} \\
\checkmark & \checkmark & \checkmark & \checkmark & \checkmark & 90.82 & \underline{81.58} & \textbf{83.49} & \underline{75.80} \\
\hline
\end{tabular}}
\end{center}
\end{table}

\begin{table}[!t]
\renewcommand{\arraystretch}{1.10}
\setlength{\tabcolsep}{4pt}
\begin{center}
\caption{Reasoning quality of SFT vs.\ MDPO, on the \texttt{<evidence\_binding>} and \texttt{<perception>} blocks of correctly predicted samples. $\uparrow$/$\downarrow$: higher/lower better.}
\label{tab:reason_quality}
\begin{tabular}{llcccc}
\hline
\textbf{Module} & \textbf{Metric} & \textbf{Dir.} & \textbf{SFT} & \textbf{MDPO} & \textbf{$\Delta$} \\
\hline
\multirow{3}{*}{Evidence}
  & NLI Coherence~\cite{he2021deberta}    & $\uparrow$ & 38.73 & \textbf{41.07} & {\color{red}+2.34} \\
  & Redundancy~\cite{golovneva2023roscoe} & $\downarrow$ & 0.35  & \textbf{0.22}  & {\color{red}-0.13} \\
  & ROUGE-L~\cite{lin2004rouge}           & $\uparrow$ & 42.99 & \textbf{43.56} & {\color{red}+0.57} \\
\hline
\multirow{2}{*}{Perception}
  & Distinct-2~\cite{li2016diversity}     & $\uparrow$ & 8.99  & \textbf{9.21}  & {\color{red}+0.22} \\
  & Distinct-3~\cite{li2016diversity}     & $\uparrow$ & 18.99 & \textbf{19.31} & {\color{red}+0.32} \\
\hline
\end{tabular}
\end{center}
\end{table}

\subsection{Ablation Studies}
All ablations report mAP under the hard $0/1$ protocol. EFR generates fine-grained labels autoregressively as text instead of scoring a fixed class set, so no continuous per-class confidence is available. Token log probabilities are only an indirect surrogate, since a single label spans several tokens whose distributions have to be recombined. We therefore use the log-probability variant only in Table~\ref{tab1}, where comparability with prior work requires it.

\noindent\textbf{Ablation on MDA.}
Table~\ref{tab:ablation_method} builds MDA over an SFT\,+\,GRPO baseline. The baseline shares a single scalar advantage across all reward heads, so gradients from easily satisfied heads dominate and mAP remains at 78.88. Task-Aware Dynamic Resampling draws rollouts in proportion to per-ORM error severity and raises mAP by 1.38 points. Uniformly correct groups carry no advantage signal, and replacing them restores usable gradients. Per-ORM Normalization then lowers mAP to 77.37, 1.51 points below the baseline it extends. Normalization rescales each reward into its own range, yet the resulting advantages are still broadcast over all tokens, so the rescaled signals compete on shared parameters and partially cancel. Token-Level Routing delivers each advantage only to the tokens that produced the corresponding output, recovering 4.21 points of mAP and reaching the best IoU$_m$ and F1 in the table. Normalization and routing therefore form one mechanism rather than two independent gains. Separating reward scales helps only when the separated signals stay separate in the backward pass.

\noindent\textbf{Ablation on Reward Components.} \label{sec:ablation_reward}
Table~\ref{tab:ablation_reward} shows that detection and attribution respond to different rewards. ACC varies by 0.06 points across all five configurations, so the binary decision is settled once $R_{\text{acc}}$ is active and no grounding reward alters it. Attribution behaves in the opposite way. With $R_{\text{fmt}}$ and $R_{\text{acc}}$ alone the model attains its highest ACC together with the weakest mAP and IoU$_m$ in the table, having learned to name a manipulation without locating it. Supervising location repairs attribution even though it adds no label information, with $R_{\text{bbox}}$ alone raising mAP by 4.14 points and $R_{\text{text}}$ alone by 3.87. Being required to point at the evidence constrains which label the model can justify. The two grounding rewards are not interchangeable. Applied in isolation, $R_{\text{text}}$ pushes IoU$_m$ below the configuration without any grounding reward, so optimizing text localization alone draws capacity away from face localization, and enabling both rewards removes the interference and yields the best mAP and F1. Adding $R_{\text{consist}}$ finally gives the best IoU$_m$ while leaving mAP and F1 unchanged to within 0.1 point, tightening anchor-evidence agreement at no measurable cost, and this configuration is retained as the default.

\begin{figure*}[t]
    \centering
    \includegraphics[width=\textwidth]{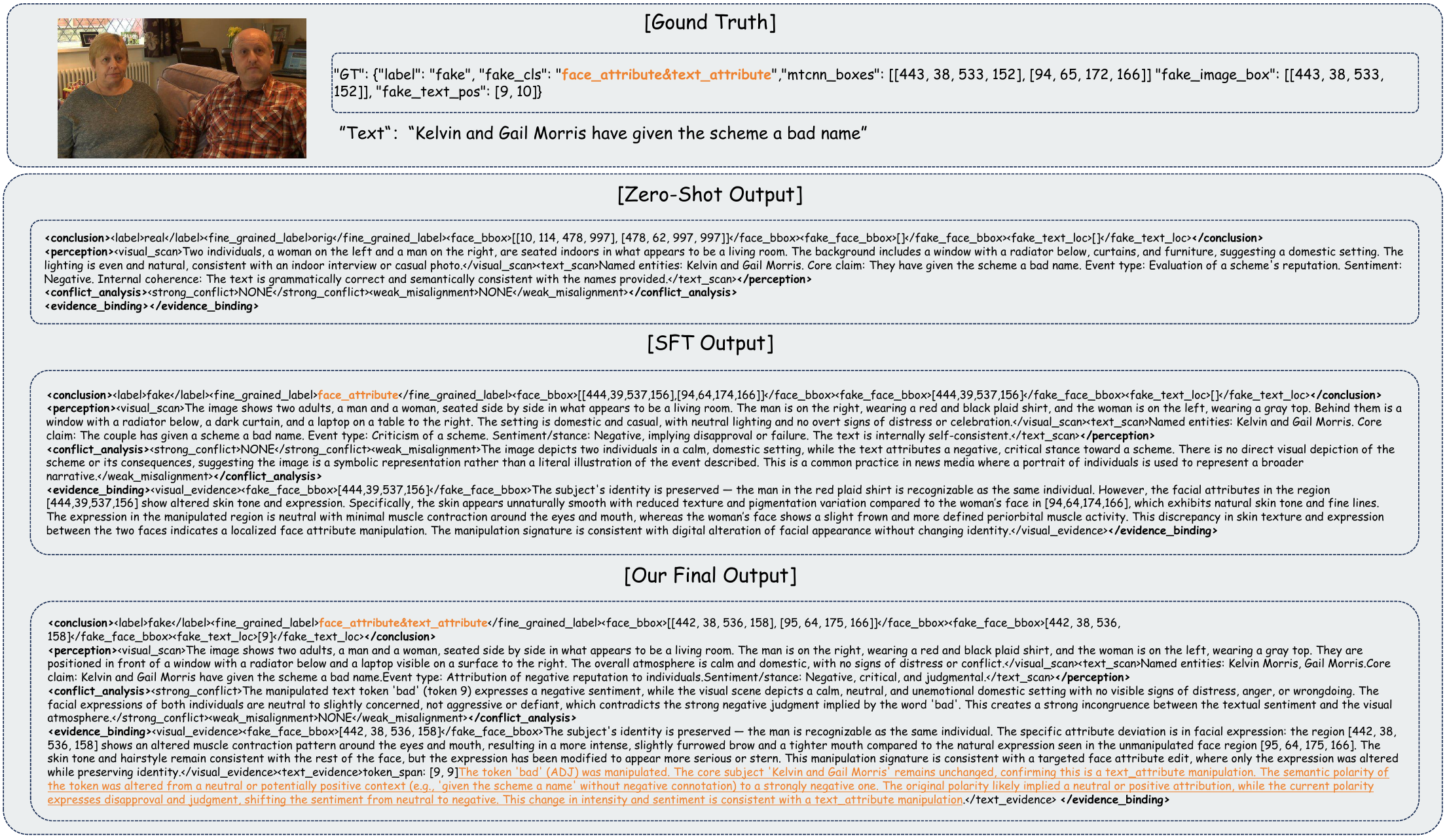}
    \caption{Forensic reasoning output of EFR on a compound \texttt{face\_attribute \& text\_attribute} sample. Zero-shot collapses to an authentic prediction; SFT recovers the face label but misses the text manipulation; EFR identifies both, localizes the altered face and token, and produces modality-specific evidence grounding for each.}
    \label{fig:vis1}
\end{figure*}

\subsection{Analysis of Generated Reasoning}
\noindent\textbf{Reasoning Quality Analysis.}
To assess the reasoning itself independently of label accuracy, we restrict the evaluation to correctly predicted manipulated samples and score the \texttt{<evidence\_binding>} sub-block of the ground-truth modality and the \texttt{<perception>} block with NLI entailment~\cite{he2021deberta}, repeated-4-gram redundancy~\cite{golovneva2023roscoe}, Distinct-2/3~\cite{li2016diversity}, and ROUGE-L~\cite{lin2004rouge}. MDPO improves every metric in Table~\ref{tab:reason_quality}. NLI entailment gains the most ($+$2.34), indicating that the cited evidence follows more tightly from the anchor, and redundancy falls by over a third. Distinct-2/3 both rise, so the shorter chains do not collapse onto a template, a common outcome of reward optimization. ROUGE-L is the only reference-based metric and is biased toward SFT, whose outputs closely follow the surface form of the reference chains. MDPO improves it nonetheless.

\noindent\textbf{Qualitative Analysis.}
Figure~\ref{fig:vis1} shows a representative compound \texttt{face\_attribute \& text\_attribute} sample under three conditions.\textcolor{orange}{Orange} marks the labels, localization, and reasoning evidence that the weaker baselines miss or predict incorrectly. The zero-shot model collapses to an \texttt{authentic} verdict with empty localization and no evidence binding, failing to engage the forensic task at all. SFT cold-start recovers the \texttt{face\_attribute} label and localizes the manipulated face, but overlooks the co-occurring text manipulation entirely, leaving the fine-grained label and text span incomplete, a typical failure mode on cross-modal compound forgeries. In contrast, EFR identifies both manipulation types, localizes the altered face region and the tampered token (``bad'', token~9) simultaneously, and grounds each conclusion in modality-specific evidence: a visual-evidence block that ties the face anchor to a concrete attribute deviation, and a text-evidence block that ties the token anchor to the sentiment shift it induces. Crucially, every cited box and span matches the coordinates declared in the conclusion, illustrating the coordinate-level anchor-evidence consistency that EFR enforces.

\section{Conclusion}
\label{sec:conclusion}

In this work, we presented EFR, an evidence-grounded forensic reasoning framework for multi-modal manipulation detection. Existing DGM4 methods offer no rationale for their predictions, while MLLM-based approaches generate explanations that remain disconnected from the predicted locations and suffer credit misassignment when a uniform signal optimizes heterogeneous heads. EFR resolves both with two coordinated components. The Anchor-and-Verify reasoning chain treats conclusion coordinates as explicit spatial anchors that downstream evidence must match, enforced by a five-component verifiable reward system during training. The Modality-Decoupled Advantage routing mechanism then delivers only task-relevant signals to each prediction head. Trained on a curated 50K forensic reasoning dataset, EFR attains state-of-the-art detection and competitive grounding on DGM4, producing reasoning records whose cited evidence is spatially consistent with the predicted anchors. We hope this perspective can inspire future work on interpretable multi-modal forensic reasoning.

\begin{acks}
This work was supported in part by the New Generation Artificial Intelligence-National Science and Technology Major Project (No. 2025ZD0123501), Chinese National Natural Science Foundation Projects U23B2054, 62276254.
\end{acks}
\bibliographystyle{ACM-Reference-Format}
\bibliography{sample-base} 
\end{document}